\documentclass[a4paper]{article}

\usepackage{INTERSPEECH2022}
\usepackage{amsmath,graphicx}
\usepackage{amsfonts}
\usepackage{multirow}
\usepackage{tablefootnote}
\usepackage{float}
\usepackage{epstopdf}
\usepackage[font=footnotesize,skip=0pt]{caption}
\usepackage{cite}
\def \figpath {./figures/}

\DeclareMathOperator*{\argmax}{arg\,max}

\title{On joint training with interfaces for spoken language understanding}

\name{
\parbox{0.8\textwidth}{\centering{
Anirudh Raju$^*$\thanks{$^*$Equal Contribution}, 
  Milind Rao$^*$,
  Gautam Tiwari,
  Pranav Dheram,
  Bryan Anderson, \\
  Zhe Zhang, 
  Chul Lee,
  Bach Bui, 
  Ariya Rastrow
  }  }  \vspace{-0.2cm}}
\address{
Amazon Alexa AI, USA
}
 \email{ranirudh@amazon.com, milinrao@amazon.com, tgautam@amazon.com}

\begin{document}
\ninept
\maketitle

\begin{abstract}
Spoken language understanding (SLU) systems extract both text transcripts and semantics associated with intents and slots from input speech utterances. SLU systems usually consist of (1) an automatic speech recognition (ASR) module, (2) an interface module that exposes relevant outputs from ASR, and (3) a natural language understanding (NLU) module. Interfaces in SLU systems carry information on text transcriptions or richer information like neural embeddings from ASR to NLU. In this paper, we study how interfaces affect joint-training for spoken language understanding. Most notably, we obtain the state-of-the-art results on the publicly available 50-hr SLURP \cite{bastianelli2020slurp} dataset. We first leverage large-size pretrained ASR and NLU models that are connected by a text interface, and then jointly train both models via a sequence loss function. For scenarios where pretrained models are not utilized, the best results are obtained through a joint sequence loss training using richer neural interfaces. Finally, we show the overall diminishing impact of leveraging pretrained models with increased training data size.

\end{abstract}


\vspace{-0.3cm}
\section{Introduction}
\vspace{-0.2cm}

Spoken dialog systems enable voice-based human-machine interactions. A key component of any spoken dialog system is its spoken language understanding (SLU) system that extracts semantic information associated with intents and named-entities from the user's speech utterances. The task of SLU systems is modeled through two separate subtasks, namely ASR and NLU. ASR generates text transcripts from input speech while NLU extracts semantic information from text transcripts. NLU, in turn, performs two subtasks, namely intent determination and slot filling. In the conventional SLU setting, ASR and NLU components are built and deployed independent of each other. They are sequentially executed with NLU consuming text transcript outputs from ASR. An overall trend towards end-to-end neural architectures is allowing on-device deployment with model sizes tuned to comply with different hardware constraints in addition to a tighter coupling between ASR and NLU. 

End-to-end speech recognition models like attention-based listen-attend-spell (LAS) \cite{chan16}, transformers \cite{zhang2020transformer}
and RNN-Transducer (RNNT) \cite{graves2012sequence} have been shown to outperform traditional RNN-HMM hybrid ASR systems, especially when trained on large speech corpora \cite{chiu2018state}. For real-time ASR systems, streaming compatible RNNT is the most suitable model choice \cite{sainath2020streaming}. Recurrent neural network \cite{mesnil2014using}
, recursive neural network \cite{guo2014joint}, and transformer \cite{chen2019bert} based NLU models have been shown to be effective for multi-task intent classification and named entity recognition. 

The easiest way of designing a full SLU sytem is to run ASR and NLU modules in sequence.  Even though such a design approach is simple to implement and practical, such pipelined SLU systems are prone to a potential downstream propagation of ASR errors or an overall SLU performance degradation as each task is trained completely independent of its upstream or downstream model. This motivated the emergence of end-to-end SLU models that directly extract serialized semantic information from speech inputs\cite{haghani2018audio,rao2020speech,lugosch2019speech,qian2021speech, rao2021mean,rongali2020exploring}. 
While ASR transcripts have been typically considered as the standard interface choice in such end-to-end SLU systems, other interfaces like word confusion networks \cite{hakkani2006beyond}, lattices \cite{huang2019adapting}, or the n-best hypotheses \cite{li2020multi} have been recently proposed.  

Since the inception of end-to-end SLU models, several attempts have been made to improve the overall SLU performance or overcome the limitations of existing end-to-end SLU systems. For those models that do not produce ASR transcripts, pretraining with an ASR task or masked language model has shown to be beneficial \cite{lugosch2019speech,qian2021speech}. Integrating NLU intent signals into an ASR system was studied in \cite{ray2021listen}. Joint or multi-task training of ASR and NLU has shown to improve the overall SLU performance in \cite{haghani2018audio,rao2020speech}. Neural interfaces (e.g. decoder hidden layer interfaces) to better facilitate the joint training of ASR and NLU were introduced in \cite{rao2020speech}. To design more streaming friendly ASR, an RNNT based SLU system to directly predict serialized semantics from audio was proposed in \cite{thomas2021rnn}.  

A variety of loss functions have been used for the training of different ASR, NLU and SLU models. For instance, cross-entropy or RNNT loss functions have been used in these models even though their metrics of interest are word error rate (WER) or semantic error rate (SemER). The REINFORCE framework in \cite{williams1992simple} enables model training in a way that the probabilities of hypotheses are boosted if they perform well on arbitrary chosen metrics. Sequence-discriminative criteria such as minimum word, phone error or minimum Bayes risk were used for ASR training in \cite{vesely2013sequence}. Motivated by mWER-based ASR training, using non-differentiable semantic criteria to directly optimize SLU metrics for training was shown to be effective in \cite{prabhavalkar2018minimum, guo2020efficient, rao2021mean, huang2022mtl}.

\vspace{-0.4cm}
\subsection{Contributions}
\vspace{-0.1cm}

While several joint training methods for SLU have been proposed in the past, it is not always straightforward to decide which interfaces are suitable for different types of ASR and NLU models. Our work is the first of its kind to fill in this research gap in this field since we study the effectiveness of interfaces, as well as the impact of pre-training, during SLU joint training. Our main contributions in this paper are:

\textbf{Joint Training with Interfaces:} We observe in various experiments that when joint training for SLU is performed with well designed interfaces, ASR and NLU metrics can be significantly improved, ranging from 5\% to 15\%. Notably, we obtain the state-of-the-art results on the SLURP dataset by using pretrained ASR and NLU models and then jointly training via a text interface.  

\textbf{Novel Neural Interfaces}: Most prior SLU models are based on simple text interfaces, with the exception of  \cite{rao2020speech,rao2021mean, seo2021integration} that use neural interfaces, but only limited to encoder-decoder ASR models. In this paper, we propose a broader set of new neural 
interfaces that are particularly suitable for transducer ASR models like RNNT and Transformer-Transducer \cite{zhang2020transformer}.   

\textbf{Effectiveness of Pretraining}: For small datasets like SLURP, we observe that pretrained NLU models have a strong impact on SLU performance. In contrast, for large datasets, we observe that pretrained NLU models do not show as much significant impact.  

\begin{figure}
    \centering
    \includegraphics[width=\columnwidth]{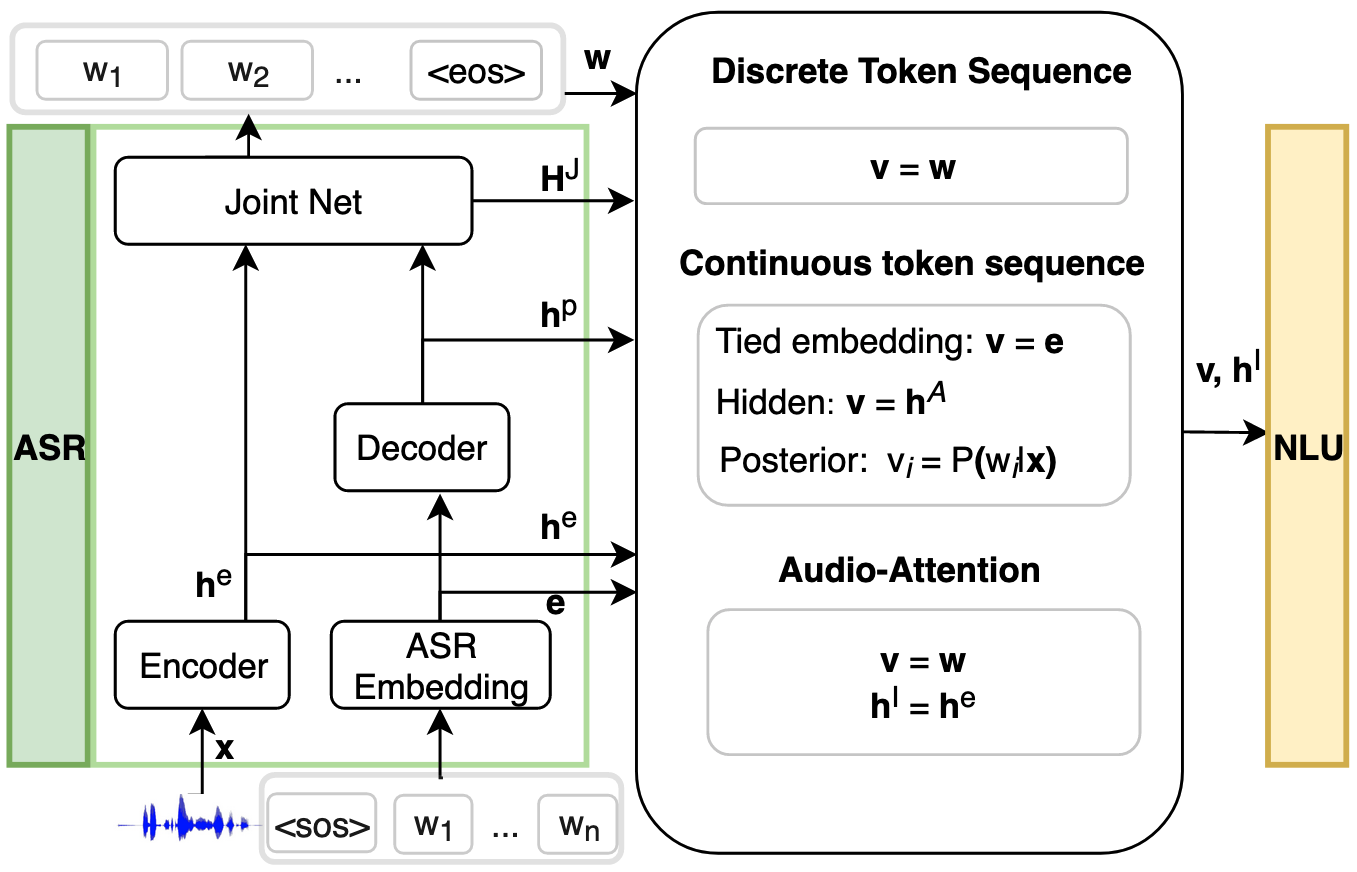}
    \caption{A SLU pipeline comprising of ASR (e.g. RNNT), blackbox NLU and different interface types to connect ASR and NLU for joint training. \label{fig:slu-pipeline}}
\end{figure}

\vspace{-0.3cm}
\section{Methods}
\label{sec:methods}

\noindent \textbf{Notation}: The input audio feature sequence is $\textbf{x} = (x_{1}, ..., x_{T})$, $x_{i} \in \mathbb{R}^{|X|\times1}$.
%
The ground truth transcript of $N$ words is $\textbf{y}=(y_{1},...,y_{N})$ where $y_{i} \in \mathcal{Y}$, the discrete set of words in the vocabulary. The ground truth NLU labels include a slot tag for each ground truth word, i.e. the discrete slot label sequence $\textbf{y}^{slot} = (y^{slot}_{1}, ..., y^{slot}_{N})$, and a label for the utterance's intent, i.e. $y^{int}$. 

\vspace{-0.3cm}
\subsection{Formulation of ASR/NLU Models and their Interfaces}
\label{sec:models}

\noindent \textbf{ASR}: E2E ASR models decode audio sequence \textbf{x} and produce a $U$ length output token sequence $\textbf{w}=(w_{1}, ..., w_{U})$ that corresponds to the best decoding hypothesis, where $w_{u} \in \mathcal{W}$, the set of discrete output labels of subword units. We consider two types of models that have been widely used for E2E ASR. 

\textbf{RNN-Transducer}: 
RNNT \cite{graves2012sequence} is a streaming compatible model that consists of three networks. An RNN encoder, analogous to an acoustic model, maps input audio feature sequence $\textbf{x}$ to hidden representations $\textbf{h}^{e} = (h^{e}_{1}, ..., h^{e}_{T} )$, $h^{e}_{t} \in \mathbb{R}^{|E|\times1}$. The prediction network takes as input the previous output label prediction $w_{u-1}$, maps to a token embedding $e_{i}$, and corresponding hidden representations $h^{p}_{u} \in \mathbb{R}^{|P|\times1}$. The joint network, typically a feed-forward neural network \cite{graves2013speech}, takes the encoder representation $h^{e}_{t}$ and the prediction network representation $h^{p}_{u}$ as input, produces intermediate joint hidden layer representations for each ${t, u}$ pair $h^{J}_{t,u} \in \mathbb{R}^{|J|\times1}$, along with corresponding logits and softmax normalized probabilities. The intermediate joint hidden layer outputs for all audio and token inputs can be represented as a rank-3 tensor $\textbf{H}^{J} \equiv h^{J}_{t,u}$,  $\textbf{H}^{J} \in \mathbb{R}^{T \times U \times |J|}$.

\textbf{Listen-Attend-Spell (LAS)}: LAS \cite{chan16} is a non-streaming model that consists of a recurrent encoder, and a recurrent decoder using an attention mechanism. The encoder maps input audio feature sequence $\textbf{x}$ to hidden representations $\textbf{h}^{e} = (h^{e}_{1}, ..., h^{e}_{T} )$. The decoder uses an attention mechanism \cite{bahdanau2016neural} to attend to the encoder representations, producing decoder hidden representations $\textbf{h}^{d} = (h_{1}^{d},...,h_{U}^{d})$, and output token probabilities $P(w_{i}|\textbf{x})$.

\noindent \textbf{ASR-NLU Interface for SLU}:
The role of an interface for SLU is to process ASR model outputs along with intermediate representations, and then subsequently produce inputs that are compatible with the NLU model. Formally, the interface output described in Eqn \ref{eqn:interface} comprises the pair of (1) discrete or continuous feature, $\textbf{v} = (v_{1}, ..., v_{U})$, of length $U$ (2) additional hidden representations, $\textbf{h}^{I}$, as an option.

\begin{equation}
\label{eqn:interface}
{\textbf{v}, \textbf{h}^{I}}  =  Interface (\textbf{H}_{asr}, \textbf{w})
\end{equation}
where $\textbf{H}_{asr}$ refers to exposed hidden representations from ASR. For RNNT, this comprises the set of encoder, prediction and joint network hidden representations: $\textbf{H}_{asr} = \{\textbf{h}^{e}, \textbf{h}^{p}, \textbf{H}^{J}\}$. For LAS, this includes the encoder and decoder final hidden layer representations: $\textbf{H}_{asr} = \{\textbf{h}^{e}, \textbf{h}^{d}\}$. Interfaces are further elaborated in Sec \ref{sec:interfaces}. \\

\noindent\textbf{NLU}:
A neural NLU model predicts the true slot label sequence of $\textbf{y}^{slot}$ and intent label $y^{int}$. The model takes as input $\textbf{v}, \textbf{h}^{I}$ from the given interface. The discrete or continuous sequence of features \textbf{v} is utilized as inputs to transformer NLU (TNLU) model with multiple layers of self-attention.
The layers can be initialized using pretrained models like BERT\cite{devlin2018bert}. When an interface produces additional representations $\textbf{h}^{I}$ in addition to \textbf{v}, NLU can attend to these representations. A transformer-decoder \cite{vaswani2017attention} is suitable for this purpose, and consists of stacked layers, each with a multi-head attention block to perform self-attention over $\mathbf{v}$ (i.e. query, key, value is $\mathbf{v}$), and a multi-head attention block that cross-attends $\textbf{h}^{I}$.

\vspace{-0.3cm}
\subsection{Interfaces and Joint Training}
\label{sec:interfaces}
We introduce various interfaces below, and summarize them in Table \ref{tab:methods}. Fig \ref{fig:slu-pipeline} depicts these interfaces in detail. All interfaces except the text interface in Sec \ref{seq:discrete-tok-interface} will be labeled as ``neural'' for the rest of the paper since these are all neural model based. 

\begin{table}
\centering
\scriptsize
\caption{Summary of various ASR - NLU interfaces. All interfaces can leverage pretrained ASR models}

\begin{tabular}{ | p{1.7cm} | p{1.6cm}  | p{1cm}  | p{2cm} | }
\hline
\label{tab:methods}
\textbf{Interface} & \textbf{Joint Training}  &  \textbf{Pretrained NLU}  & \textbf{Interface Type}  \\ \hline
Text &  Seq-loss &  Y  & Discrete token seq \\ \hline
Tied embeddings & MLE + Seq-loss   &  N &  \multirow{3}{*}{Continuous token seq} \\ 
Posterior & MLE + Seq-loss    &  Y &  \\ 
Hidden  & MLE  + Seq-loss  &  N &  \\ \hline
\multirow{2}{*}{Audio-Attention} & MLE  + Seq-loss   & \multirow{2}{*}{N}  & \multirow{2}{2.5cm}{Discrete token seq + Audio representations} \\ 
&    &  &   \\ \hline

\end{tabular}
\end{table}

\vspace{-0.3cm}
\subsubsection{Discrete Token Sequence: Text Interface}
\label{seq:discrete-tok-interface}

This is mainly suitable for conventional SLU systems, in which ASR produces one-best text transcripts consumed by NLU. That is, ASR's one-best label sequence \textbf{w} of discrete tokens is an input to NLU, namely $\textbf{v} = \textbf{w}$. $\textbf{h}^{I} = null$, i.e. optional additional representations are not emitted by these interfaces. The two models can be decoupled, have potentially different vocabularies, and be trained independent of each other. 

\vspace{-0.3cm}
\subsubsection{Continuous Token Sequence Interface}
\label{seq:continuous-tok-interface}
Continuous token sequence interfaces produce an output \textbf{v} of length $U$, where each output $v_{i}$ is a continuous representation corresponding to each token in ASR's one-best hypothesis.  $\textbf{h}^{I} = null$ i.e. optional additional representations are not emitted by these interfaces.

\begin{figure}[h]
\footnotesize
\centering
    \includegraphics[width=5cm]{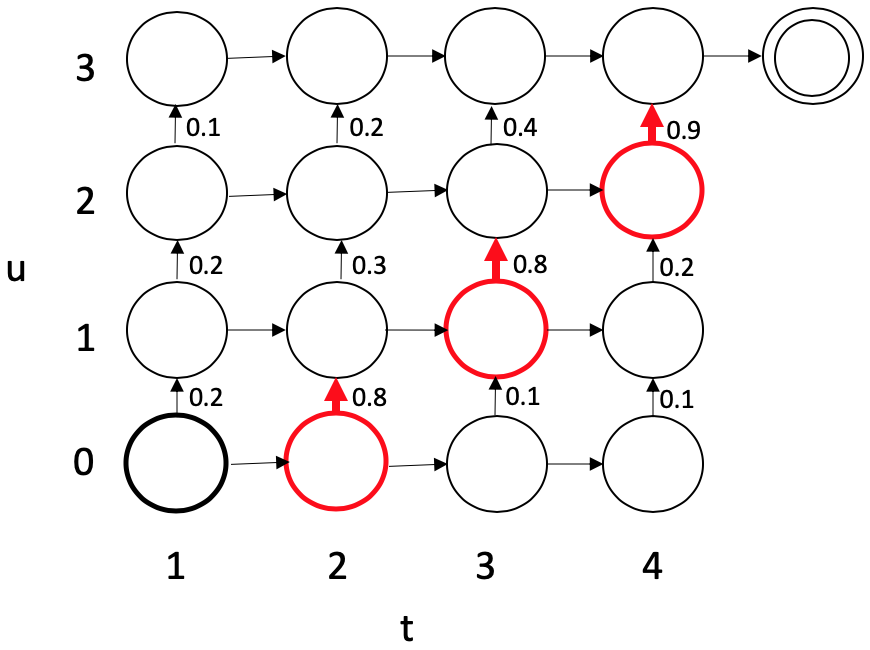}
    \caption{\footnotesize Each node represents the state $(t, u)$ with the bottom left node $(1,0)$ being the start state. Vertical transitions correspond to transitions with probability $P(w_{u}  | t, u-1)$. Horizontal transitions correspond to blank transitions \cite{graves2012sequence} with probability $P(\phi  | t, u-1)$. Each label $w_{u}$ has a node with the maximum probability vertical transition (red) at row $u-1$, marked in red. The hidden states from red nodes form $ \textbf{h}^{A}$, which is input to neural NLU.}
    \label{fig:rnnt-max-prob}
\end{figure}

\noindent \textbf{Tied Embedding Interface}: The input token embedding matrix from ASR, $Emb_{asr}$, that is available from the LAS decoder or RNNT prediction network processes the discrete ASR's one-best token sequence \textbf{w} to produce token embeddings $\textbf{e} = (e_{1},...,e_{U})$. Effectively, this ties the input token embedding params across ASR and NLU.
$\textbf{v} = \textbf{e} = Emb_{asr}(\textbf{w})$. \linebreak
\noindent \textbf{Posterior Interface}: Posterior probabilities corresponding to the one-best token sequence \textbf{w} of ASR are passed as inputs to NLU. Each $v_{i} \in \mathbb{R}^{|V|\times1}$ is a continuous representation corresponding to the posterior probability of the token over vocabulary $|V|$, i.e. $v_{i} = P(w_{i}|\textbf{x})$. \linebreak
\noindent \textbf{Hidden Interface}: It produces a hidden layer representation corresponding to each token output from ASR,
$\textbf{v} = \textbf{h}^{A}$ where $\textbf{h}^{A} = (h^{A}_{1}, ..., h^{A}_{U})$. Strategies to obtain the token-aligned hidden representation sequence $\textbf{h}^{A}$ of length $U$ from ASR depend on the E2E ASR architecture. \\
For LAS-based ASR, a hidden decoder representation is available for each step of decoding producing a token output. Similar to that of \cite{rao2020speech}, the decoder hidden sequence $\textbf{h}^{d}$ of length $U$ can be directly utilized, i.e. $\textbf{h}^{A} = \textbf{h}^{d}$.
\noindent For RNNT-based ASR, designing a hidden interface is not as straightforward since the prior work \cite{rao2020speech} on hidden interfaces for LAS cannot be extended to RNNT as easily. Note that the hidden output representations from the joint network corresponding to a decoding hypothesis in RNNT are rank-3 tensors $\textbf{H}^{J}$ (i.e. $h^{J}_{t,u} \in \mathbb{R}^{|J|\times1}$). \linebreak In our paper, we propose a novel way of processing these joint network hidden representations and then providing e a $U$ length input to NLU. This hidden interface maps intermediate hidden outputs $h^{J}_{t,u}$ corresponding to the final joint network layer to $ \textbf{h}^{A} = (h^{A}_{1}, ..., h^{A}_{U})$. For each subword output $w_{u}$, we pick a joint hidden representation $h^{A}_{u}$ at the input frame index $i_{u}$ that has the maximum label transition probability from the previous subword $w_{u-1}$ corresponding to state $(i_{u}, u-1)$. We now have $h^{A}_{u} = h^{J}_{t=i_{u},u}$ where $i_{u} = \argmax_t P(w_{u}  | t, u-1)$ as shown in Fig \ref{fig:rnnt-max-prob}. Although this hidden interface does not explicitly enforce monotonicity in time (i.e. frame indices) for the chosen hidden representations, we observe in various experiments that training results in chosen hidden representations which are monotonic in time. The intuition behind this interface is in Fig \ref{fig:rnnt-max-prob}, illustrating how the joint hidden outputs corresponding to each label are selected.

\vspace{-0.2cm}
\subsubsection{Token Sequence + Audio Representations Interface}
\label{seq:discrete-tok-audio-interface}
\vspace{-0.2cm}
\noindent \textbf{Audio-Attention Interface}: 
This novel interface exposes additional audio representations from ASR besides $\textbf{v}$. TNLU can cross-attend \cite{vaswani2017attention} to these, to have an access to richer information, i.e. $\textbf{v} = \textbf{w}; \textbf{h}^{I} = \textbf{h}^{e}$. Unlike other prior interfaces described in our paper, this interface adds trainable weights to NLU due to its cross-attention.

\subsection{Performance Metrics}
\label{sec:perf-metrics}
\noindent \textbf{Word Error Rate (WER)}: ASR metric that is defined as the normalized minimum word edit distance. Sentence ER (SER) is the percentage of utterances with word errors.

\noindent \textbf{Intent Classification Error Rate (ICER) and Intent Accuracy (IntAcc)}: An SLU metric, ICER refers to the percentage (\%) of utterances whose intent predictions by the model are incorrect. IntAcc is obtained as 100\%-ICER.

\noindent \textbf{Semantic Error Rate (SemER)}: Introduced in \cite{rao2021mean}, for this SLU metric, all slots (including intents) in the reference are compared to the given hypothesis and marked as (1) Substitution: if the slot name is correct but not the value (2) Insertion: if the slot is added in hypothesis or (3) Deletion: if the slot is missing in hypothesis. SemER refers to the number of slot errors normalized by the number of reference slots.

\noindent \textbf{SLU-F1}: Introduced in  \cite{bastianelli2020slurp}, this SLU metric combines span based $F\-1$ score in named entity recognition with a text based distance measure to accommodate ASR errors. For each reference slot: (1) True Positive (TP) if the slot name and the value match, (2) False Negative (FN) if it is a slot deletion error, and (3) False Positive (FP) if it is a slot insertion error. A slot substitution error counts as TP with penalty that equals to word and character error rates of the slot value, adding to FN and FP values. SLU-F1 refers to the $F\-1$ score of these.

\vspace{-0.3cm}
\subsection{Loss Functions for Joint Training}
\vspace{-0.1cm}
\subsubsection{MLE Loss}
\label{sec:mle-loss}
\noindent \textbf{ASR Loss}: Given an input audio sequence, the negative log posterior of ASR's output label sequence is the ASR loss, defined as $L_{asr} = -ln P(\textbf{w}|\mathbf{x})$. For RNNT, the loss function and its gradient are calculated using a forward-backward algorithm \cite{graves2012sequence}.

\noindent \textbf{NLU Loss}: The cross-entropy loss function is employed in both the output slot distribution of each token and the utterance intent. These are summed to obtain $L_{nlu}$.

\noindent \textbf{Multi-Task Loss}: Neural interface based SLUs can be jointly trained using both ASR and NLU parameters via a multi-task loss function as $L_{mt} = L_{asr} +  L_{nlu}.$ 

\vspace{-0.3cm}
\subsubsection{Sequence Loss}
\vspace{-0.1cm}

This loss function directly optimizes the expected SLU performance metrics (e.g. WER or SLU-F1) over the output candidate distribution produced. For the non-differentiable SLU error metric $M(C, c^{\star})$ of a SLU output candidate $C$ and its corresponding ground truth annotation $c^{\star}$, its expected metric cost can be approximated by assuming that the probability mass is concentrated in the top n-hypotheses $\bar{\mathcal{C}}$, similar to previous sequence discriminative training approaches in \cite{vesely2013sequence, prabhavalkar2018minimum, rao2021mean, guo2020efficient}. The gradient of the model weights $\theta$ is computed with respect to the candidate probabilities $\bar{p}(c; \theta)$ normalized over the n-best hypotheses as shown in Eqn \ref{eq:loss-seq-grad}.  
\vspace {-0.2cm}

\vspace{-0.2cm}

\begin{align}
\label{eq:loss-seq-grad}
\nabla_\theta  L_{seq} = \nabla \mathbb{E}[M(C, c^{\star})] &\approx \sum_{c\in\bar{\mathcal{C}}} M(c, c^\star) \nabla_\theta \bar{p}(c; \theta) 
\end{align}

\begin{figure*}[t]
\scriptsize
\begin{minipage}[t]{0.55\textwidth}
\begin{table}[H]
\centering
\scriptsize
\caption{\footnotesize Performance numbers on the public SLURP speech dataset. Our results are shown below prior work as well as NLU results. 
\label{tab:slurp_performance}}. 

\begin{tabular}{|p{3.5cm}|p{1cm}|l|l|l|}

\hline
 \textbf{Model} & \textbf{Interface} & \textbf{WER} & \textbf{Intent Acc} & \textbf{SLU-F1} \\ \hline
 \multicolumn{5}{|c|}{NLU models with ground-truth text input} \\ \hline
  RoBERTa \cite{seo2021integration} & - & - & 87.73 & 84.34 \\ 
  TNLU & - & - & 85.80 & 75.25  \\  
  BertNLU & - &- & 88.14 & 85.97 \\ \hline \hline
\multicolumn{5}{|c|}{SLU models with speech input} \\ \hline 
Baseline \cite{bastianelli2020slurp} & text & 16.3 & 78.33 & 70.84 \\
Transformer ASR - RoBERTa \cite{seo2021integration} & posterior & 16.7 & 82.93 & 71.20 \\ \hline
RNNT $\rightarrow$ TNLU (independently trained) & text & 16.9 & 78.00 &  67.56 \\ \hline
\multirow{3}{*}{RNNT $\rightarrow$ TNLU (jointly trained)} & text & 16.7 & 78.71 & 67.89 \\     
   & hidden & 16.9 & 78.79 & 68.37 \\ 
   & attention & 16.9 & 79.50 & 68.67 \\ \hline 
 RNNT $\rightarrow$ BertNLU (independently trained) &  text  & 16.9 & 82.00 & 71.28  \\ 
 RNNT $\rightarrow$ BertNLU (jointly trained) & text & \textbf{15.2} & 82.45 & \textbf{72.35} \\ \hline
  \multirow{2}{*}{LAS $\rightarrow$ TNLU (jointly trained)} & text & 17.5 & 78.57 & 67.93 \\
 & hidden & 17.4 & 79.50 & 69.35 \\ \hline
 LAS $\rightarrow$ BertNLU (jointly trained) & text & 16.4 & 82.30 & 70.65 \\ \hline
 
\end{tabular}
\end{table}
\end{minipage}\hspace{0.02\textwidth}%
\begin{minipage}[t]{0.4\textwidth}
\centering

\begin{figure}[H]
\centering
\includegraphics[width=0.9\linewidth]{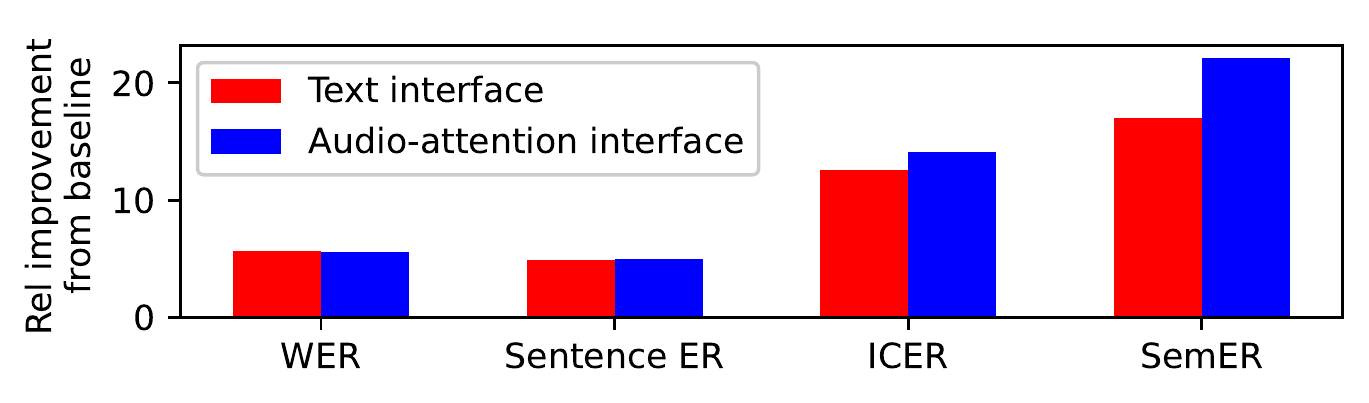}
\caption{\footnotesize  Impact of joint training with text and neural interfaces vs independently trained RNNT, NLU baseline.}
\label{fig:textneural}
\end{figure}
\vspace{-0.1cm}
\begin{figure}[H]
\centering
\includegraphics[width=0.9\textwidth]{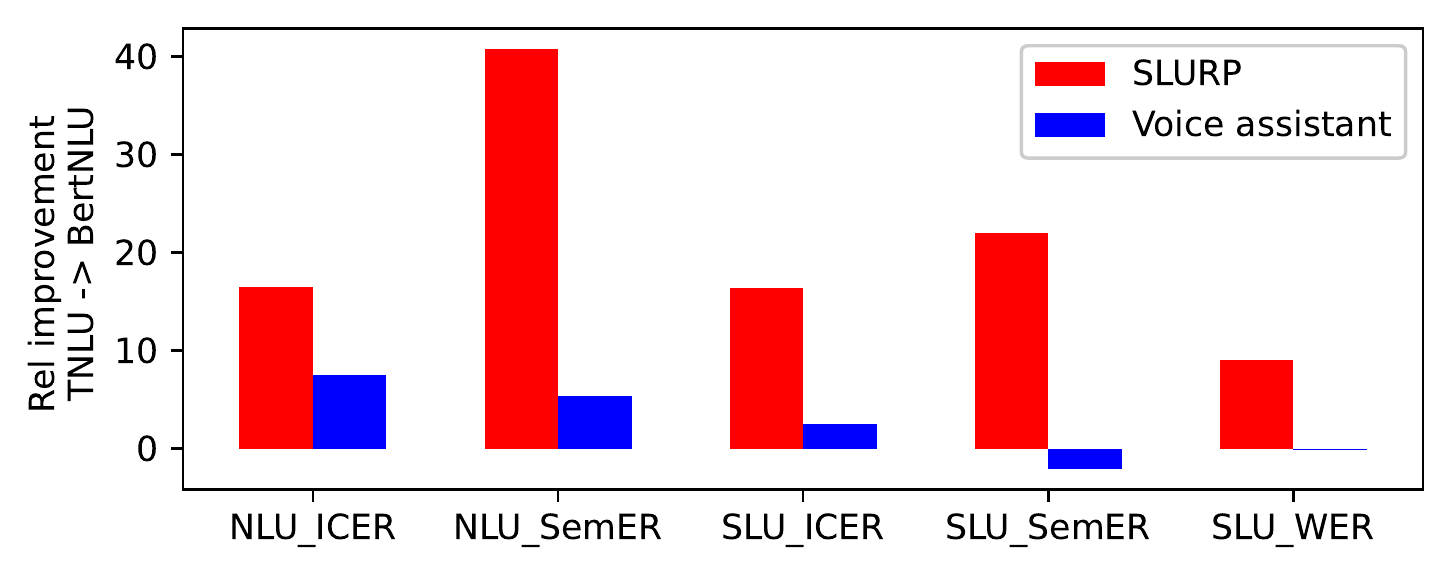}
\caption{\footnotesize Measuring the impact of pretraining with dataset size. NLU\_ metrics refer to ground-truth text-input, while SLU\_ metrics refer to speech input}
\label{fig:pretrainimp}
\end{figure}
\end{minipage}
\vspace{-0.3cm}
\end{figure*}

\vspace{-0.6cm}
\section{Experimental Setup}
\label{sec:expts}
\vspace{-0.2cm}

\subsection{Datasets}
We perform our experiments on the following datasets that include parallel speech transcriptions as well as NLU annotations. \\
\textbf{\textit{SLURP Dataset}}: A public dataset \cite{bastianelli2020slurp} that consists of 58-hrs of speech data with 72k utterances representing 18 domains (scenarios), 46 intents (actions), and 56 slots (entities). We do not use any additional provided synthetic data \\
\textbf{\textit{Voice Assistant Dataset}}: A dataset that consists of 19K-hrs of de-identified in-house far-field speech data. These are English utterances that are directed to voice assistants, representing 26 domains, 176 intents and 141 slots. The evaluation set has 68-hrs of data. As a baseline ASR dataset for comparison, we use another 23k-hrs corpus with speech transcriptions only to pretrain our LAS and RNNT. 

\vspace{-0.3cm}
\subsection{Model details}
\textbf{Features}: Audio features are 64-dim log-mel filterbank energies computed over a 25ms window with 10ms shifts; stacked and downsampled to a 30ms frame rate. Ground truth text is tokenized into subword tokens using a unigram language model of vocabulary of 2500 (RNNT), 4500 (LAS), and 30000 (pretrained-BERT) \cite{kudo2018subword} . \\
\noindent \textbf{ASR/NLU Models}: RNNT has 68M params with a 5x1024 encoder, 2x1024 prediction network, output projection to 512, a 1x512 feedforward joint network with tanh activation \cite{graves2013speech}. LAS has 77M params with a 5x512 BiLSTM encoder, 2x1024 decoder, and 4-head 768-unit attention. ASR decoding uses beam width of 4. TNLU has 20M params with 8-head 3x1024 self-attention units. The per-token representations are passed through a dense layer to get the per-token slot logits, and are maxpooled and passed through a 2-layer feed-forward network with 512 units for the intent and domain detection. The transformer-decoder is 8-head 2x512 units for cross-attention adding 8M params. The BertNLU models make use of the pretrained BERT model \cite{devlin2018bert,wolf2019huggingface} with 111M params with 12-head 12x768 self-attention layers, and extra layers as described in TNLU for the slot, intent, and domain detection. \\
\textbf{Training}: Independently trained ASR, NLU models use respective MLE loss as in Sec \ref{sec:mle-loss}. Jointly trained models use sequence loss or multi-task MLE + seq-loss, as in Table \ref{tab:methods}.

\vspace{-0.3cm}
\section{Results and Discussion}
\label{sec:results}
\vspace{-0.2cm}

We reports our experimental results for \textit{SLURP} in Table \ref{tab:slurp_performance} and those for \textit{Voice Assistant} in Figures \ref{fig:textneural} and \ref{fig:pretrainimp}.

\vspace{-0.3cm}
\subsection{Joint Training with Interfaces Improves ASR and NLU}
\vspace{-0.1cm}

First, in Table \ref{tab:slurp_performance}, we observe that the SLU metrics like SLU-F1 are degraded for all the SLU models with speech input, when compared to the NLU models with ground-truth text input. This is not surprising since it is well known that the performance of downstream NLU depends on the performance of upstream ASR in pipelined SLU systems. As a consequence, the robustness of NLU to ASR errors is important in the design of any SLU system.  A common observation through almost all joint training results in Table~\ref{tab:slurp_performance} and Figure~\ref{fig:textneural} is that
regardless of either text interface or other interfaces, joint training via sequence loss or MLE loss significantly improves the performance of ASR and NLU by 5-15\%, compared to the baseline SLU model with independently trained ASR and NLU. More importantly, in Table \ref{tab:slurp_performance} row ``RNNT$\rightarrow$ BertNLU (jointly trained)'', we obtain the state-of-the-art results for SLURP by using pre-trained ASR/NLU and then jointly training using the text interface. Moreover, the joint training of ASR with pretrained NLU produces a significant ASR improvement for SLURP.

\vspace{-0.3cm}
\subsection{Neural Interfaces Improve NLU}
\vspace{-0.1cm}
Our proposed continuous token sequence and attention interfaces let NLU have an easier access to acoustic context and ASR confusion information. We validate this by observing a modest yet statistically significant performance improvement of SLU metrics in Table~\ref{tab:slurp_performance}  for the jointly trained RNNT$\rightarrow$ TNLU and LAS$\rightarrow$ TNLU when these use the attention interface and the hidden interface respectively. We further validate this with the larger \textit{Voice Assistant} dataset. In Figure \ref{fig:textneural}, we observe that joint training with the attention interface yields a substantial SemER improvement of 22$\%$, while with the text interface we obtain a 16$\%$ SemER improvement, compared to the independently trained baseline. 

\vspace{-0.4cm}
\subsection{Impact of pretraining}
\vspace{-0.1cm}
The SLURP dataset is only 58 hrs and the training data does not fully capture acoustic or semantic variations. We observe an outsized impact of pretraining NLU models based on a BERT language model. In Figure \ref{fig:pretrainimp}, on the SLURP dataset, we see that switching from a TNLU model trained from scratch to the BERT based pretrained model leads to improvements of 15-40\% in NLU metrics for the NLU model and 10-20\% improvements in SLU metrics. However, on the much larger \textit{Voice Assistant} dataset, the NLU models display smaller differences with and without pretraining and this leads to near-identical performance of the SLU model.

\vspace{-0.4cm}
\section{Conclusions and Future Work}
\vspace{-0.2cm}

In this work, we studied the impact of jointly training an ASR model and an NLU model that are connected with well-designed interfaces. We obtained state-of-the-art results on the public SLURP dataset by leveraging pretrained ASR, NLU models that are connected by a text interface and jointly trained via a sequence loss function. Moreover, we proposed richer neural interfaces that show best performance when pretrained models are not utilized, and studied the diminishing impact of pretraining based on training data size. As future work, we plan to extend our neural interfaces to work for pretrained NLUs.

\vspace{1mm}
\noindent\footnotesize{\textbf{Acknowledgments:} We thank Thejaswi, Kai, Jing, Kanthashree, Samridhi, Ross, Nathan, Andreas, Jasha and Zhiqi for helpful discussions.}


\bibliographystyle{IEEEtran}

\bibliography{mybib}

\end{document}